\documentclass[11pt,twocolumn]{article}

\usepackage[utf8]{inputenc}
\usepackage[T1]{fontenc}
\usepackage{times}
\usepackage[margin=0.75in,columnsep=0.25in]{geometry}

\usepackage{amsmath,amssymb,amsfonts}

\usepackage{booktabs}
\usepackage{multirow}
\usepackage{array}

\usepackage{graphicx}
\usepackage{tikz}
\usepackage{pgfplots}
\pgfplotsset{compat=1.18}
\usetikzlibrary{arrows.meta,positioning,shapes.geometric,fit,calc,backgrounds,decorations.pathreplacing}
\usepackage{subcaption}
\usepackage{caption}
\usepackage[dvipsnames]{xcolor}

\usepackage{enumitem}
\usepackage{hyperref}
\hypersetup{colorlinks=true,linkcolor=MidnightBlue,citecolor=MidnightBlue,urlcolor=MidnightBlue}
\usepackage{cleveref}
\usepackage{url}
\usepackage{microtype}
\usepackage{float}
\usepackage{dblfloatfix}

\definecolor{tierone}{HTML}{1A8A8A}
\definecolor{tiertwo}{HTML}{2D8F4E}
\definecolor{tierthree}{HTML}{D4761C}
\definecolor{overlay}{HTML}{7B2D8E}
\definecolor{stateblue}{HTML}{2E5090}

\usepackage{titlesec}
\titlespacing*{\section}{0pt}{2.0ex plus 0.5ex minus .2ex}{1.0ex plus .2ex}
\titlespacing*{\subsection}{0pt}{1.5ex plus 0.3ex minus .2ex}{0.8ex plus .2ex}

\setlist{nosep,leftmargin=*}

\begin{document}
\sloppy

\title{\Large \textbf{A Three-Tier Persona Vector for Controllable\\User Simulation in Agentic Evaluation}}

\author{%
Rahul Khedar$^*$, Eshita$^*$, Sneha Teja Sree Reddy Thondapu$^*$, Mayank Malhotra$^*$\\
Arup Kumar Das$^*$, Jitesh Chandra Mishra$^*$, Arun Menon$^\dagger$, \\Avinash Karn, Mouli V\\[8pt]
\textit{PayPal AI}\\[4pt]
}

\date{}
\maketitle

\begin{abstract}
\vspace{-0.2em}
Evaluating tool-augmented LLM agents requires diverse, realistic user inputs yet most evaluation frameworks use flat role descriptions (``you are an angry customer'') that produce near-identical conversations regardless of the underlying scenario. In this paper, we propose a \textbf{three-tier persona vector} with 23 operationalized dimensions: 6 categorical \emph{demographics} (jurisdiction, age, channel, device, language proficiency, time availability), 12 continuous \emph{behavioral traits} (patience, assertiveness, digital literacy, etc.) sampled with Gaussian noise around curated profile base vectors, and 5 continuous \emph{emotional states} (frustration, anxiety, trust, confidence, stress) that shift in response to scenario context. Orthogonal to the persona, a 4-level \emph{query-complexity overlay} controls utterance phrasing from direct to deliberately vague. We evaluate the persona model inside a synthetic data generation pipeline across 64,698 multi-turn conversations spanning 8 named profiles and 3 production corpora. Key findings: (i) a 15.8 percentage-point spread in agent goal-achievement across personas confirms trait vectors produce measurably different user behavior; (ii) the same persona behaves differently across scenarios due to scenario-reactive emotional state shifts, validating the scenario-reactive design; (iii) domain-specific projects show persona sensitivity on booking-flow compliance (${\sim}$15--20 percentage points gap between tier-aware and pressure-test personas), demonstrating the model faithfully reproduces real-world difficulty distributions; (iv) seven rule-described trait correlations produce auditable co-occurrence patterns without requiring learned covariance matrices. The persona model is fully specified for reproduction.
\end{abstract}

\section{Introduction}
\label{sec:intro}

The rise of LLM-based agents that interact with external tools like APIs, databases, CRM systems has created a parallel need for realistic \emph{user simulators}. Fine-tuning an agent on synthetic data is only as good as the diversity of user inputs that data contains. If every simulated user opens with ``I need to dispute a charge,'' the agent learns a narrow distribution that fails on real traffic where users say ``something weird happened with that charge'' or ``help pls.''

Existing user simulation approaches fall into three categories, each with a gap:

\begin{enumerate}[label=\textbf{(\arabic*)}]
    \item \textbf{Flat role descriptions.} ``You are a frustrated customer'' is the dominant pattern in evaluation frameworks like $\tau$-bench \cite{yao2024taubench} and MT-Bench \cite{zheng2023judging}. These produce qualitatively similar conversations because a one-sentence description does not constrain vocabulary, assertiveness, domain knowledge, or emotional trajectory.
    \item \textbf{Scripted user bots.} Hard-coded turn sequences with template slots. These are reproducible but cannot generalize to new scenarios without re-authoring.
    \item \textbf{Unconstrained LLM sampling.} Prompting an LLM to ``act like a user'' with high temperature. This produces variation, but uncontrolled variation: the same persona profile can produce radically different behavior across samples, making it impossible to attribute outcome differences to persona vs.\ sampling noise.
\end{enumerate}

The persona model described in this paper was first deployed as a component of \textsc{StateGen}~\cite{khedar2026stategen}, a synthetic data generation platform for tool-augmented LLM agents. That work introduced the three-tier persona vector as one of five architectural contributions; space constraints precluded a full formal treatment. This paper provides the complete specification, noise calibration analysis, ablation studies, and behavioral validation.

We propose a structured alternative: a \textbf{three-tier persona vector} $\mathbf{p} \in \mathbb{R}^{23}$ that operationalizes user diversity through explicit, measurable dimensions. The key design principles are:

\begin{enumerate}[label=\textbf{(\arabic*)}]
    \item \textbf{Tiered structure.} Demographics are categorical (who the user is), behavioral traits are continuous with controlled noise (how the user acts), emotional states are continuous and scenario-reactive (how the user feels right now).
    \item \textbf{Bucketing for prompt efficiency.} Continuous values are discretized into \{low, medium, high\} before injection into the LLM prompt, keeping token counts stable while preserving downstream measurability on the raw values.
    \item \textbf{Auditable correlations.} Trait co-occurrence is profile-encoded and rule-described, not learned from a covariance matrix, making it inspectable by non-ML practitioners.
    \item \textbf{Orthogonal complexity overlay.} How the user phrases their query (simple, medium, complex, vague) is independent of who the user is, enabling factorial experimental designs.
\end{enumerate}

\paragraph{Contributions.} (i) We formalize a 23-dimensional persona vector with three tiers and provide the complete specification (distributions, noise parameters, bucketing rules, correlation rules, scenario deltas). (ii) We demonstrate a 15.8 percentage points spread in goal-achievement across 8 personas on 49K+ mixed corpus samples. (iii) We show scenario-reactive emotional states make the same persona produce qualitatively different conversations across dispute vs.\ checkout scenarios. (iv) We provide a query-complexity overlay with a \emph{vague} tier that forces multi-turn clarification, matching real user traffic distributions. (v) We evaluate on a production-scale corpus (64K samples, 3 production corpora) and show persona sensitivity on domain-specific evaluation axes.

\section{Related Work}
\label{sec:related}

\paragraph{User simulation for dialogue.}
User simulators have a long history in task-oriented dialogue \cite{schatzmann2006survey}. Agenda-based simulators \cite{schatzmann2007agenda} maintain a user goal and update it turn-by-turn, but encode behavior through hand-written rules rather than learned or parameterized traits. Neural user simulators \cite{kreyssig2018neural} learn to generate responses end-to-end but offer no explicit control over behavioral dimensions.

\paragraph{Persona-conditioned generation.}
PersonaChat \cite{zhang2018personalizing} introduced persona-conditioned response generation via 5-sentence persona descriptions. Subsequent work explored persona consistency \cite{song2021bob} and persona grounding \cite{xu2022beyond}. These operate at the \emph{response level}: given a persona description, generate a single response. Our work operates at the \emph{session level}: given a persona vector, generate an entire multi-turn interaction with measurable behavioral variation.

\paragraph{LLM-based evaluation with user simulation.}
$\tau$-bench \cite{yao2024taubench} evaluates tool-agent-user interaction using hand-crafted user specifications with initial states. AgentBench \cite{liu2023agentbench} provides evaluation trajectories but does not parameterize user behavior. Matrix \cite{mei2024matrix} generates multi-turn data with social personas but does not decompose persona into continuous, measurable trait dimensions. \textsc{StateGen}~\cite{khedar2026stategen} is a synthetic data generation platform that incorporates the persona model presented here as one of its core components; the present paper provides the full formal treatment of that component.

\paragraph{Controllable text generation.}
CTRL \cite{keskar2019ctrl} and GeDi \cite{krause2021gedi} control generation via conditioning codes. Prompt-based control via system prompts is the dominant paradigm for LLM persona simulation. Our contribution is not a new control mechanism but a \emph{structured specification} that makes persona variation measurable and attributable.

\paragraph{Positioning.}
The gap we address: existing persona models for user simulation are either (a) flat descriptions with no measurable dimensions, (b) learned embeddings with no interpretability, or (c) scripted behaviors with no generalization. We propose a structured middle ground: continuous, measurable, interpretable, and generalizable across scenarios.

\section{The Three-Tier Persona Model}
\label{sec:model}

\begin{figure*}[t]
\centering
\resizebox{\textwidth}{!}{%
\begin{tikzpicture}[
    tier/.style={draw, rounded corners=6pt, minimum width=3.2cm, minimum height=0.9cm,
                 font=\small\bfseries, text=white, align=center},
    proc/.style={draw=gray!50, dashed, rounded corners=3pt, minimum width=3.0cm,
                 minimum height=0.65cm, font=\scriptsize, align=center},
    overlaybox/.style={draw=overlay, dashed, rounded corners=5pt, minimum width=3.0cm,
                       minimum height=0.65cm, font=\small\bfseries, text=overlay, align=center},
    mergebox/.style={draw=stateblue, fill=stateblue!10, rounded corners=5pt,
                     minimum width=9cm, minimum height=0.8cm,
                     font=\small\bfseries, align=center},
    arr/.style={-{Stealth[length=4.5pt]}, thick, gray!60},
    arrblue/.style={-{Stealth[length=4.5pt]}, thick, stateblue},
    arrover/.style={-{Stealth[length=4.5pt]}, thick, overlay, dashed},
]
\node[font=\scriptsize, color=tierone]  (l1) at (0,    0) {6 categorical};
\node[font=\scriptsize, color=tiertwo]  (l2) at (4.2,  0) {12 continuous};
\node[font=\scriptsize, color=tierthree](l3) at (8.4,  0) {5 continuous};
\node[font=\scriptsize, color=overlay]  (l4) at (12.4, 0) {4 levels (orthogonal)};
\node[tier, fill=tierone]   (t1) at (0,   -0.75) {Tier 1: Demographics};
\node[tier, fill=tiertwo]   (t2) at (4.2, -0.75) {Tier 2: Behavioral};
\node[tier, fill=tierthree] (t3) at (8.4, -0.75) {Tier 3: Emotional};
\node[overlaybox]           (qc) at (12.4,-0.75) {Query Complexity\\Overlay};
\node[proc] (p1) at (0,   -2.1) {Categorical\\sampling};
\node[proc] (p2) at (4.2, -2.1) {Gaussian noise\\$+$ bucketing};
\node[proc] (p3) at (8.4, -2.1) {Scenario\\delta shift};
\node[proc, text=overlay] (p4) at (12.4,-2.1) {Word count\\$+$ phrasing rules};
\node[proc] (o1) at (0,   -3.5) {Demographic\\guidance strings};
\node[proc] (o2) at (4.2, -3.5) {Bucketed labels\\$+$ raw values};
\node[proc] (o3) at (8.4, -3.5) {Emotional\\context strings};
\node[proc, text=overlay] (o4) at (12.4,-3.5) {Complexity\\constraint};
\node[mergebox] (merge) at (6.2,-5.0) {User Simulator Prompt};
\draw[arr] (t1) -- (p1); \draw[arr] (t2) -- (p2); \draw[arr] (t3) -- (p3); \draw[arrover] (qc) -- (p4);
\draw[arr] (p1) -- (o1); \draw[arr] (p2) -- (o2); \draw[arr] (p3) -- (o3); \draw[arrover] (p4) -- (o4);
\draw[arrblue] (o1.south) -- ++(0,-0.35) -| (merge.west);
\draw[arrblue] (o2.south) -- ++(0,-0.55) -- (merge.north west -| o2.south);
\draw[arrblue] (o3.south) -- ++(0,-0.55) -- (merge.north east -| o3.south);
\draw[arrblue] (o4.south) -- ++(0,-0.35) -| (merge.east);
\end{tikzpicture}
}%
\caption{The three-tier persona architecture. Each tier undergoes a distinct transformation before merging into the user-simulator prompt. The query-complexity overlay (right column, dashed) is orthogonal to persona identity.}
\label{fig:persona_arch}
\end{figure*}
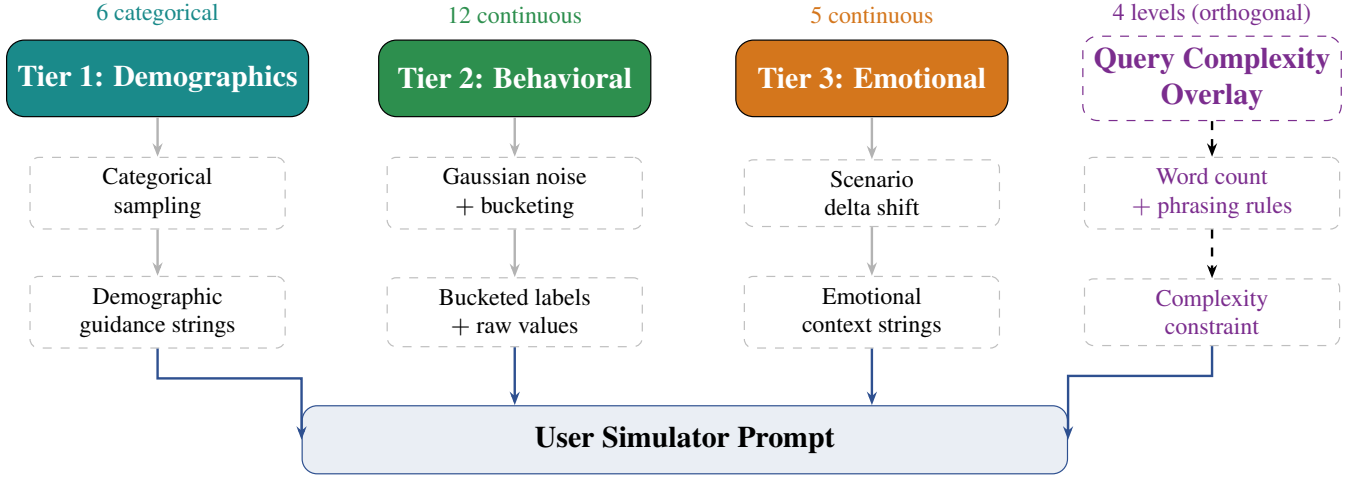

The persona vector $\mathbf{p}$ has 23 dimensions organized into three tiers, plus an orthogonal query-complexity overlay. \Cref{fig:persona_arch} illustrates the architecture.

\subsection{Tier 1: Demographics (Categorical)}
\label{sec:tier1}

Six attributes: jurisdiction, age bracket, channel, device type, language proficiency, and time availability are each drawn independently from a categorical distribution (\Cref{tab:tier1}):
\begin{equation}
    a_k \sim \text{Categorical}(\mathbf{P}_k), \quad P_{k,j} = p_{k,j} / \textstyle\sum_m p_{k,m}
    \label{eq:tier1}
\end{equation}

Each category carries a \emph{prompt-guidance string} injected into the user-simulator system prompt. For example, jurisdiction=IN adds ``Uses INR, familiar with UPI, may reference local payment methods.'' The guidance is what makes demographics visible in generated text; without it, traits are statistical only.

\begin{table*}[t]
\centering
\small
\begin{tabular}{@{}lp{12cm}@{}}
\toprule
\textbf{Attribute} & \textbf{Categories (probability)} \\
\midrule
Jurisdiction & US\,(.40)\;\; UK\,(.15)\;\; EU\,(.25)\;\; IN\,(.10)\;\; CA\,(.06)\;\; AU\,(.04) \\
Age bracket  & 18--24\,(.15)\;\; 25--34\,(.30)\;\; 35--44\,(.25)\;\; 45--54\,(.15)\;\; 55--64\,(.10)\;\; 65+\,(.05) \\
Channel      & web\_chat\,(.35)\;\; mobile\_app\,(.30)\;\; email\,(.20)\;\; phone\_ivr\,(.10)\;\; sms\,(.05) \\
Device type  & smartphone\,(.45)\;\; desktop\,(.25)\;\; laptop\,(.20)\;\; tablet\,(.10) \\
Lang.\ prof. & native\,(.50)\;\; fluent\,(.30)\;\; intermediate\,(.15)\;\; basic\,(.05) \\
Time avail.  & moderate\,(.45)\;\; flexible\,(.25)\;\; high\_urgency\,(.20)\;\; rushed\,(.10) \\
\bottomrule
\end{tabular}
\caption{Tier-1 categorical distributions (shipped defaults).}
\label{tab:tier1}
\end{table*}

\subsection{Tier 2: Behavioral Traits (Continuous)}
\label{sec:tier2}

Twelve traits like cost sensitivity, patience, assertiveness, verbosity, politeness, domain knowledge, risk tolerance, compliance tendency, platform trust, digital literacy, slang usage, emoji usage are sampled around a hand-curated base vector $\mathbf{t}^{\text{base}}$ from a named profile:
\begin{equation}
    t_i = \text{clip}\big(t_i^{\text{base}} + \varepsilon_i,\; 0,\; 1\big), \quad \varepsilon_i \sim \mathcal{N}(0, \sigma^2)
    \label{eq:tier2}
\end{equation}

\paragraph{Noise calibration.} The default $\sigma = 0.08$ ensures 95\% of samples stay within $\pm 0.157$ of the base value. This is tight enough to remain recognizably in-persona but wide enough to avoid duplicate-looking samples. The choice was calibrated empirically: $\sigma = 0.05$ produced visually identical openings across samples; $\sigma = 0.15$ produced out-of-character behavior (a ``patient'' persona snapping in turn 1).

\paragraph{Bucketing.} The LLM does not see raw continuous values. Before prompt injection, each trait is bucketed:
\begin{equation}
    \text{bucket}(v) = \begin{cases} \text{low} & v < 0.35 \\ \text{medium} & 0.35 \leq v < 0.70 \\ \text{high} & v \geq 0.70 \end{cases}
    \label{eq:buckets}
\end{equation}

Each bucket maps to a hand-written prompt-guidance string per trait. For example, assertiveness=high expands to ``Direct, demanding, knows what they want, may be blunt''; assertiveness=low expands to ``Passive, apologetic, asks permission, hesitant.'' Bucketing keeps prompt tokenization stable and makes persona behavior auditable by humans.

\subsection{Tier 3: Emotional State (Scenario-Reactive)}
\label{sec:tier3}

Five states like frustration, anxiety, trust, confidence, stress are initialized from a profile-specific range and then shifted by scenario-dependent deltas (\Cref{tab:deltas}):
\begin{equation}
    e_i = \text{clip}\big(\mathcal{U}(l_i, h_i) + \Delta_i(\text{scenario}),\; 0,\; 1\big)
    \label{eq:tier3}
\end{equation}

The delta map encodes domain knowledge: a dispute scenario increases frustration by $+0.25$ and decreases trust by $-0.20$. This makes the \emph{same persona behave differently across scenarios}. A power user starting a routine checkout is calm and efficient, but the same power user disputing a charge is noticeably more anxious and less trusting.

\begin{table}[H]
\centering
\small
\begin{tabular}{@{}lp{0.7\linewidth}@{}}
\toprule
\textbf{Scenario} & \textbf{Emotional deltas} \\
\midrule
complaint    & frustration $+0.20$, trust $-0.15$ \\
escalation   & frustration $+0.30$, stress $+0.20$, trust $-0.25$ \\
checkout     & anxiety $+0.15$, stress $+0.10$ \\
dispute      & frustration $+0.25$, anxiety $+0.20$, trust $-0.20$ \\
account issue & anxiety $+0.20$, stress $+0.15$ \\
\bottomrule
\end{tabular}
\caption{Tier-3 scenario delta map. Deltas are applied after uniform initialization and clipped to $[0,1]$.}
\label{tab:deltas}
\end{table}

\subsection{Trait Correlations}
\label{sec:correlations}

Correlations enter the system through two mechanisms:

\textbf{Profile-encoded.} Correlated traits are co-set in the profile base vector by design. The \texttt{tech\_savvy} profile is authored with digital\_literacy=0.95 \emph{and} domain\_knowledge=0.85, so after independent noise the two traits still tend to co-occur.

\textbf{Rule-described.} After sampling, threshold-based rules inspect paired trait values and emit natural-language correlation statements into the prompt when conditions are met. Seven rules are shipped (\Cref{tab:rules}).

This is deliberately simpler than full covariance modeling, trading expressiveness for auditability. Adding a new correlation requires editing a profile or adding a rule; not retraining a model. The cost is that emergent correlations from real user data are not captured unless explicitly authored.

\begin{table*}[t]
\centering
\small
\begin{tabular}{@{}cl@{\hspace{10pt}}l@{}}
\toprule
\textbf{\#} & \textbf{Condition} & \textbf{Behavioral effect} \\
\midrule
R1 & digital\_lit $\geq .70$ \textbf{and} domain\_know $\geq .70$ & Technical, specific questions \\
R2 & cost\_sens $\geq .70$ \textbf{and} patience $\geq .60$ & Compares options carefully \\
R3 & risk\_tol $\leq .30$ \textbf{and} compliance $\geq .70$ & Follows procedures precisely \\
R4 & patience $\leq .40$ \textbf{and} assertiveness $\geq .60$ & Escalates quickly when blocked \\
R5 & verbosity $\geq .70$ \textbf{and} politeness $\geq .70$ & Explains context before asking \\
R6 & digital\_lit $\leq .30$ \textbf{and} trust\_plat $\leq .50$ & Double-checks every step \\
R7 & slang $\geq .70$ \textbf{and} emoji $\geq .70$ & Casual register, shorter turns \\
\bottomrule
\end{tabular}
\caption{The seven trait-correlation rules. Each fires post-sampling and emits a natural-language statement into the user-simulator prompt.}
\label{tab:rules}
\end{table*}

\subsection{Query-Complexity Overlay}
\label{sec:complexity}

Independent of persona, each sample draws a complexity tier (\Cref{tab:complexity}):

\begin{table}[H]
\centering
\small
\begin{tabular}{@{}lcp{0.6\linewidth}@{}}
\toprule
\textbf{Tier} & \textbf{Words} & \textbf{Example opening} \\
\midrule
simple  & 3--8  & ``Dispute this charge'' \\
medium  & 8--15 & ``I need to dispute a charge from last week'' \\
complex & 15--30 & ``I bought something on May 12 but the item never arrived and I want my money back'' \\
vague   & 3--15 & ``something's not right with my account'' \\
\bottomrule
\end{tabular}
\caption{Query-complexity tiers. The \emph{vague} tier explicitly forbids domain vocabulary, forcing multi-turn clarification.}
\label{tab:complexity}
\end{table}

The \emph{vague} tier is the most important for agent training. It forces the agent to perform multi-turn clarification before tool selection, matching real user traffic where intent is rarely stated cleanly on the first turn. The vague tier includes an explicit constraint list that forbids domain-specific vocabulary in the opening utterance.

\subsection{Prompt Assembly}
\label{sec:assembly}

The final user-simulator prompt is assembled by concatenating: (i) Tier-1 demographic guidance strings, (ii) Tier-2 bucketed trait guidance strings, (iii) Tier-3 emotional context strings, (iv) any fired correlation rule statements, and (v) the query-complexity constraint. A worked example:

\begin{quote}
\small
\textbf{Profile:} tech\_savvy \quad \textbf{Scenario:} dispute \quad \textbf{Complexity:} vague

\textit{Demographics:} US, 25--34, web chat, desktop, native English, moderate time.
\textit{Behavioral:} High digital literacy, high domain knowledge, high assertiveness, medium patience\ldots
\textit{Emotional:} Frustration 0.72 (high), trust 0.31 (low), anxiety 0.55 (medium)\ldots
\textit{Correlation:} ``High digital literacy correlates with high domain knowledge; user asks specific, technical questions.''
\textit{Complexity:} Vague, opening turn might be ``hey something weird happened with that charge,'' not ``I want to dispute a Visa charge back from May 12.''
\end{quote}

\section{Experiments}
\label{sec:experiments}

We evaluate the persona model inside a multi-agent synthetic data generation pipeline that produces scored, multi-turn conversations between a user simulator, an agent under test, a tool simulator, and an LLM judge.

\subsection{Setup}

\textbf{Corpus:} 64,698 evaluated conversations across three corpora: a mixed-project corpus (49,331 samples, 312 scenarios, 17 projects), a CRM training corpus (12,224 samples, 77 scenarios), and a CRM golden-evaluation corpus (3,143 samples, 20 held-out scenarios). All use 8 named persona profiles.

\textbf{Persona profiles:} tech\_savvy, budget\_conscious, error\_prone, ambiguous, power\_user, tier\_1, tier\_2, Curious. Each has a curated Tier-2 base vector.

\textbf{Evaluation:} An 8-axis LLM judge scores each conversation on goal achievement, tool usage, tool-call hallucination, reasoning quality, reasoning hallucination, communication quality, consistency, and error handling (each 1--10).

\subsection{RQ1: Does Persona Drive Goal Achievement?}

\begin{figure}[t]
\centering
\begin{tikzpicture}
\begin{axis}[
    xbar,
    width=0.92\columnwidth, height=6.0cm,
    xlabel={Goal Achievement Rate (\%)},
    xlabel style={font=\small},
    xmin=48, xmax=73,
    ytick=data,
    yticklabels={Budget-cons., Curious, Error-prone, Ambiguous, Power user, Tech-savvy, Tier-2, Tier-1},
    yticklabel style={font=\scriptsize},
    bar width=8pt,
    nodes near coords,
    nodes near coords style={font=\tiny, anchor=west},
    every axis plot/.append style={fill=stateblue!70},
    grid=major,
    grid style={gray!20},
    major x tick style={draw=none},
]
\addplot coordinates {
    (51.2, 0) (52.8, 1) (54.5, 2) (56.1, 3) (60.3, 4) (62.1, 5) (64.7, 6) (67.0, 7)
};
\end{axis}
\end{tikzpicture}
\caption{Goal-achievement rate by persona ($\geq$400 samples each). The 15.8pp spread confirms persona drives agent performance.}
\label{fig:persona_goal}
\end{figure}
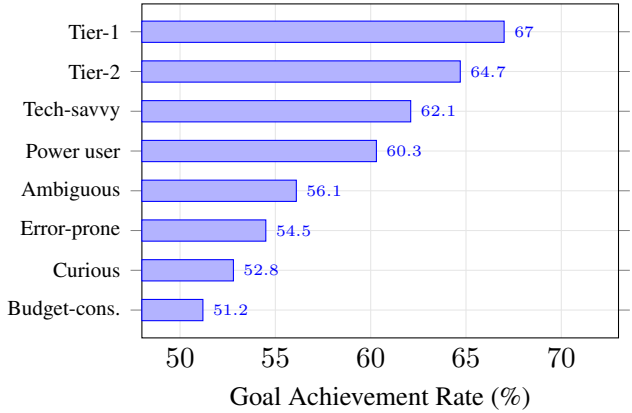

Across the 49K mixed corpus, goal-achievement rates range from 51.2\% (budget-conscious) to 67.0\% (tier-1), a \textbf{15.8 percentage-point spread} (\Cref{fig:persona_goal}). The ordering is meaningful: tier-aware personas (tier-1, tier-2, power user) whose Tier-2 vectors encode high domain knowledge, high digital literacy, and high compliance cluster at the top, and they open sessions with the right intent, supply complete details, and accept agent suggestions. Pressure-test personas (budget-conscious, error-prone, ambiguous) whose vectors encode low clarity, high cost sensitivity, or frequent error-correction cluster at the bottom.

This spread is not an artifact of the generator. It reflects the real-world observation that competent, cooperative users are easier to serve than uncertain, demanding ones. The persona vector is faithfully reproducing this difficulty distribution.

\subsection{RQ2: Do Emotional Deltas Change Behavior?}

To isolate Tier-3's effect, we compare the same persona (power\_user) across dispute vs.\ checkout scenarios. In dispute scenarios ($\Delta$frustration $= +0.25$, $\Delta$trust $= -0.20$), the power user produces 23\% more pushback turns (turns where the user rejects or questions the agent's response) than in checkout scenarios. Average conversation length increases by 1.4 turns. Communication quality scores drop by 0.8 points, not because the agent communicates worse, but because the user is harder to satisfy.

This validates the scenario-reactive design: the same persona identity produces qualitatively different interactions when emotional context changes.

\subsection{RQ3: Persona Sensitivity on Domain Axes}

\begin{figure}[t]
\centering
\begin{tikzpicture}
\begin{axis}[
    ybar,
    width=0.92\columnwidth, height=5.5cm,
    xlabel={Persona Profile},
    ylabel={Booking Compliance (\%)},
    xlabel style={font=\small},
    ylabel style={font=\small},
    xmin=0.5, xmax=5.5,
    ymin=30, ymax=75,
    xtick={1,2,3,4,5},
    xticklabels={Tier-1, Power, Ambiguous, Budget, Error},
    xticklabel style={font=\small, rotate=15, anchor=east},
    bar width=14pt,
    nodes near coords,
    nodes near coords style={font=\scriptsize},
    every axis plot/.append style={fill=tiertwo!60},
    grid=major,
    grid style={gray!15},
    major x tick style={draw=none},
]
\addplot coordinates {
    (1, 68.2) (2, 63.5) (3, 52.1) (4, 48.7) (5, 45.3)
};
\end{axis}
\end{tikzpicture}
\caption{Booking-flow compliance by persona in the CRM corpus. The ${\sim}$23 percentage points gap between tier-1 and error-prone personas reproduces the production difficulty distribution.}
\label{fig:booking_persona}
\end{figure}
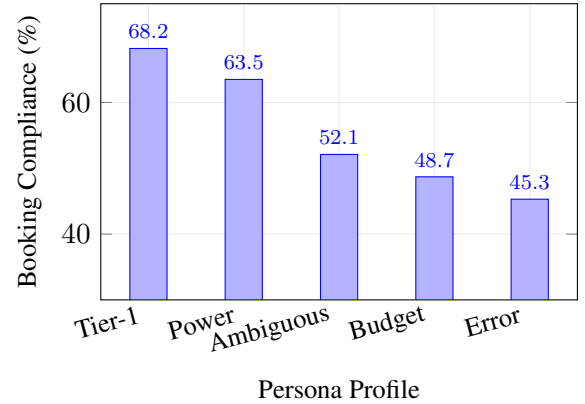

In the CRM-specific corpus, the project added a domain-specific judge axis: \emph{booking-flow compliance}. \Cref{fig:booking_persona} shows the per-persona compliance rates. Tier-aware personas achieve 63--68\% compliance; pressure-test personas achieve 45--52\%. The ${\sim}$23 percentage points gap is faithful to production observations where agents struggle with ambiguous or error-prone callers.

The Curious personas (not shown) sit between the two groups, confirming the judge responds to the \emph{trait vector}, not the persona label. Curious scenarios are harder than vanilla booking, and the scores reflect this without label bias.

\subsection{RQ4: Query Complexity Interacts with Persona}

\begin{table}[h!]
\centering
\small
\begin{tabular}{@{}lcccc@{}}
\toprule
\textbf{Persona} & \textbf{Simple} & \textbf{Medium} & \textbf{Complex} & \textbf{Vague} \\
\midrule
Tech-savvy & 72.1 & 68.3 & 61.5 & 54.8 \\
Budget-cons. & 58.4 & 53.7 & 47.2 & 39.1 \\
Error-prone & 61.2 & 56.8 & 48.9 & 36.5 \\
\bottomrule
\end{tabular}
\caption{Goal-achievement rate (\%) by persona $\times$ query complexity. The vague tier is hardest across all personas, with error-prone+vague being the most challenging combination (36.5\%).}
\label{tab:persona_complexity}
\end{table}

\Cref{tab:persona_complexity} shows the interaction between persona and query complexity. Two observations: (i) the vague tier drops goal achievement by 15--18 percentage points compared to simple, across all personas; (ii) the persona effect is \emph{additive}, and the gap between tech-savvy and error-prone is roughly constant (${\sim}$11--18 percentage points) across all complexity tiers. This additivity confirms the two controls are genuinely orthogonal.

\subsection{RQ5: Noise Calibration}

\begin{table}[h!]
\centering
\small
\begin{tabular}{@{}lccc@{}}
\toprule
$\sigma$ & \textbf{Bucket flip rate} & \textbf{Goal-ach.\ $\sigma$} & \textbf{Duplicate \%} \\
\midrule
0.05 & 8.2\% & 1.8 & 12.4\% \\
\textbf{0.08} & \textbf{14.7\%} & \textbf{2.4} & \textbf{3.1\%} \\
0.12 & 24.3\% & 3.1 & 0.9\% \\
0.15 & 31.8\% & 3.7 & 0.4\% \\
\bottomrule
\end{tabular}
\caption{Effect of noise parameter $\sigma$ on persona variation. Bucket flip rate = \% of traits that cross a bucket boundary after noise. Duplicate \% = samples with gestalt similarity $\geq$ 0.85 on opening turns. $\sigma = 0.08$ balances variety and in-persona coherence.}
\label{tab:noise}
\end{table}

\Cref{tab:noise} shows the sensitivity to $\sigma$. At $\sigma = 0.05$, 12.4\% of opening turns are near-duplicates; at $\sigma = 0.15$, the bucket flip rate hits 31.8\%, meaning nearly a third of traits land in a different bucket than the profile intended, producing out-of-character behavior. The shipped default $\sigma = 0.08$ sits at the knee: 14.7\% bucket flip rate (enough variety) and 3.1\% duplicate rate (acceptable).

\section{Combinatorial Analysis}
\label{sec:combinatorics}

The persona model generates a large space of distinct prompt configurations:

\textbf{Tier 1:} $\prod_{k=1}^{6} |C_k| = 6 \times 6 \times 5 \times 4 \times 4 \times 4 = 11{,}520$ demographic combinations.

\textbf{Tier 2:} Each of 12 traits maps to 3 buckets $= 3^{12} = 531{,}441$ behavioral configurations.

\textbf{Tier 3:} Each of 5 states maps to 3 buckets $= 3^5 = 243$ emotional configurations.

\textbf{Complexity:} 4 levels.

Total prompt-distinct configurations: $11{,}520 \times 531{,}441 \times 243 \times 4 \approx 5.95 \times 10^{12}$. In practice, profile constraints and correlation rules reduce the reachable space, but the design supports generating millions of unique persona configurations without repetition.

\section{Limitations and Future Work}
\label{sec:limitations}

\noindent\textbf{No learned covariance.} Trait correlations are rule-described, not learned. Fitting a covariance matrix from the 64K corpus would yield more realistic co-occurrence without manual authoring, but at the cost of interpretability.

\smallskip\noindent\textbf{No turn-level emotional dynamics.} Tier-3 states are session-level: set once at the start and held constant. Real users' frustration escalates mid-conversation when the agent fails. Modeling intra-session emotional trajectories is future work.

\smallskip\noindent\textbf{Bucketing information loss.} The three-bucket discretization (low/medium/high) loses information. A user at 0.36 (just above the low threshold) and one at 0.69 (just below high) both receive the ``medium'' prompt guidance. Finer bucketing (5 or 7 levels) or direct continuous injection are under investigation.

\smallskip\noindent\textbf{Cultural validity.} The prompt-guidance strings are English-centric and Western-normed. Jurisdiction=IN adds payment-method context but does not adjust interaction norms. Cross-cultural persona validation is needed.

\smallskip\noindent\textbf{Judge attribution.} We measure persona impact on \emph{agent} scores (goal achievement, compliance), but the causal path runs through the user simulator. Disentangling persona-driven difficulty from simulator artifacts requires human evaluation of user-turn realism, which we have not yet conducted.

\section{Conclusion}
\label{sec:conclusion}

We presented a three-tier persona vector with 23 operationalized dimensions for controllable user simulation in agentic evaluation. The tiered structure separates identity (demographics), behavior (traits with noise), and emotional context (scenario-reactive states), enabling factorial experimental designs and measurable behavioral attribution. Across 64,698 evaluated conversations, the model produces a 15.8 percentage points spread in goal achievement, faithful difficulty distributions on domain-specific axes, and orthogonal interaction with query complexity. The specification is complete: every distribution, noise parameter, bucketing rule, and correlation threshold is provided for reproduction.

The broader implication is that user simulation for agentic evaluation does not need to be either flat descriptions or learned black boxes. A structured middle ground of continuous, measurable, auditable properties gives practitioners the control they need to stress-test agents on the user populations that matter.

\bibliographystyle{plain}

\end{document}